\documentclass[journal]{IEEEtran}

\usepackage{verbatim}
\usepackage{amsfonts}
\usepackage{amssymb}
\usepackage{stfloats}
\usepackage{cite}
\usepackage{graphicx}
\usepackage{psfrag}
\usepackage{amsmath}
\usepackage{array}
\usepackage{epstopdf}
\usepackage{authblk}
\usepackage{graphicx} 
\usepackage{amsthm} 
\usepackage{lipsum}
\usepackage{verbatim} 
\usepackage{authblk}
\usepackage{mathtools}
\usepackage{cuted}
\usepackage{booktabs}

\usepackage{amsmath}
\usepackage{mathrsfs}

\usepackage{algorithmic}

\usepackage{array}

\ifCLASSOPTIONcompsoc
\usepackage[caption=false,font=normalsize,labelfont=sf,textfont=sf]{subfig}
\else
\usepackage[caption=false,font=footnotesize]{subfig}
\fi

\usepackage{url}
\usepackage{graphicx,amsmath,amssymb,amsfonts}
\usepackage{algorithmic,algorithm}

\usepackage{hyperref}
\hypersetup{colorlinks=true}

\usepackage{setspace}

\makeatletter
\newcommand{\rmnum}[1]{\romannumeral #1}
\newcommand{\Rmnum}[1]{\expandafter\@slowromancap\romannumeral #1@}
\makeatother

\usepackage{framed} 

\usepackage{cancel}

\usepackage{booktabs}
\usepackage{tabularx,makecell,multirow}

\usepackage[table]{xcolor}
\usepackage{pifont} 

\begin{document}
	\bstctlcite{ref:BSTcontrol}
	
	\title{Semi-Federated Learning for Collaborative Intelligence in Massive IoT Networks}
	
	\author{Wanli~Ni, Jingheng~Zheng, and~Hui~Tian
		\vspace{-5 mm}
		\thanks{This work was supported in part by the National Natural Science Foundation of China under Grant 62071068 and in part by the Beijing University of Posts and Telecommunications (BUPT) Excellent Ph.D. Students Foundation under Grant CX2022301. \textit{(Corresponding author: Hui Tian.)}}
		\thanks{Wanli~Ni, Jingheng~Zheng and~Hui~Tian are with the State Key Laboratory of Networking and Switching Technology, Beijing University of Posts and Telecommunications, Beijing 100876, China (e-mail: \{charleswall, zhengjh, tianhui\}@bupt.edu.cn).}
	}
	
	\maketitle
	
	\begin{abstract}
		Implementing existing federated learning in massive Internet of Things (IoT) networks faces critical challenges such as imbalanced and statistically heterogeneous data and device diversity.
		To this end, we propose a semi-federated learning (SemiFL) framework to provide a potential solution for the realization of intelligent IoT.
		By seamlessly integrating the centralized and federated paradigms, our SemiFL framework shows high scalability in terms of the number of IoT devices even in the presence of computing-limited sensors.
		Furthermore, compared to traditional learning approaches, the proposed SemiFL can make better use of distributed data and computing resources, due to the collaborative model training between the edge server and local devices.
		Simulation results show the effectiveness of our SemiFL framework for massive IoT networks.
		The code can be found at {\tt https://github.com/niwanli/SemiFL\_IoT}.
	\end{abstract}
	
	
	\vspace{-4 mm}
	\section{Introduction}
	\vspace{-2 mm}
	Using machine learning-driven intelligent solutions, the communication paradigm of future wireless networks is shifting from Internet of Things (IoT) to connected intelligence \cite{ Yang2020Federated}.
	Although the ubiquitous connectivity of massive IoT devices is beneficial to provide a vast amount of real-time data for various intelligent IoT services, how to realize the vision of collaborative intelligence in massive IoT networks is still an open question \cite{Ni2021Federated}.
	Specifically, by collecting data samples from massive IoT devices, edge platforms using centralized learning (CL) can utilize their powerful computational resources to train high-performance models, but this raises the problems of high communication cost and privacy leakage risk at the vulnerable wireless edge \cite{Nguyen2021Efficient}.

	Instead of sending raw data, the federated learning (FL) paradigm allows edge devices to process data locally, and then only model parameters (e.g., weights or gradients) are transmitted to the edge platform \cite{Ni2021Federated, Nguyen2021Efficient, Zhang2022Robust}.
	However, all edge devices in FL should have sufficient computing resources for local model training, which may not always be true in practice \cite{Ahmet2021Hybrid}.
	Namely, the datasets owned by these computing-limited devices are difficult to be involved into the FL model training process, which may lead to performance degradation as well as inefficient resource utilization \cite{Ni2022Integrating, Ahmet2021Hybrid, Huang2022Wireless}.
	To this end, we propose a novel SemiFL framework having two computing layers, which reaps the benefits of both CL and FL by taking into account both data uploading and gradient updating.
	
	\begin{figure}[tp]
		\setlength{\abovecaptionskip}{-0.5 mm}
		\centering
		\includegraphics[width= 3.2 in]{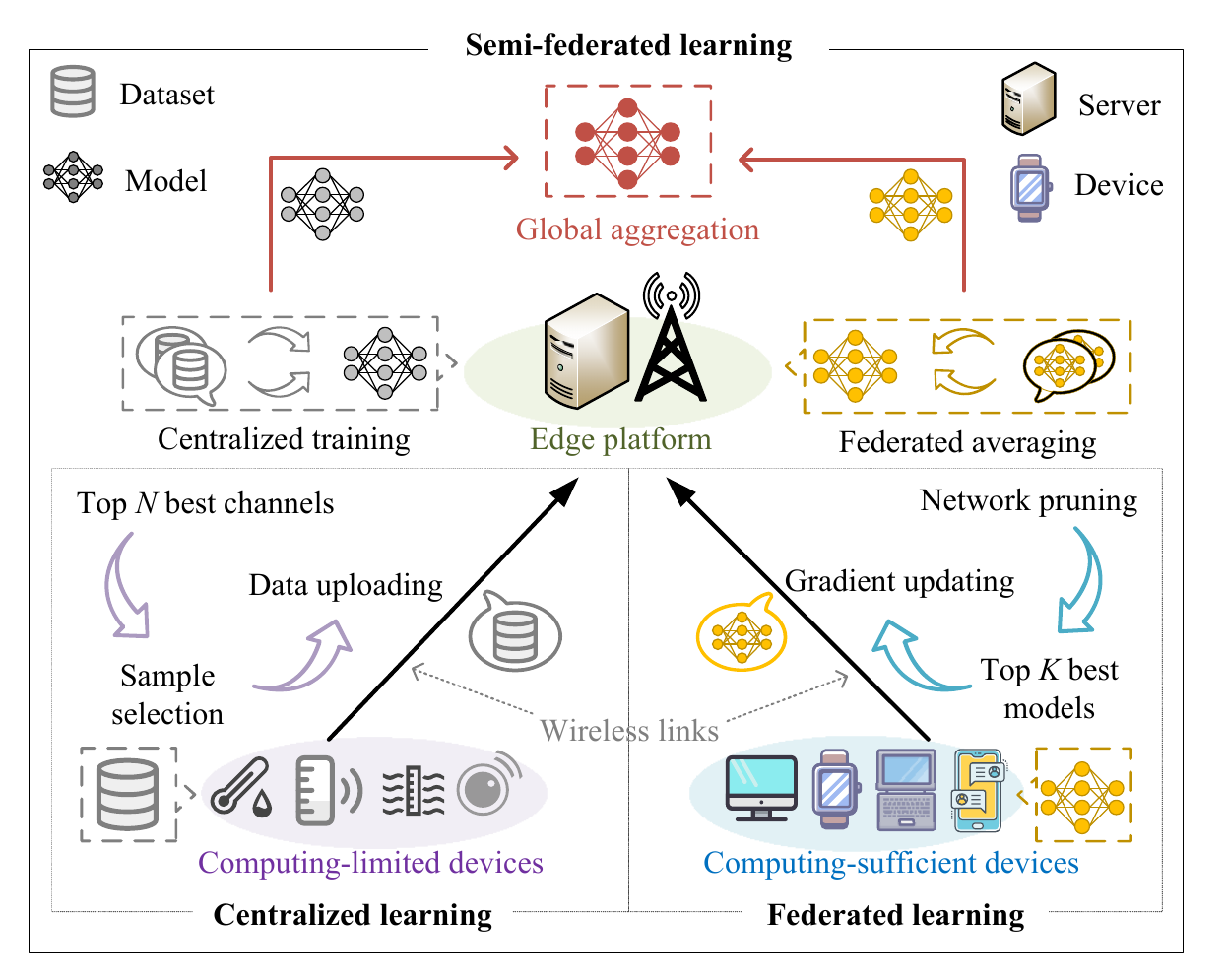}
		\caption{An illustration of the proposed SemiFL framework. The edge platform consists of a base station and an edge computing server. The computing-limited devices are designated as CL users to upload raw data for centralized training, while other computing-sufficient devices are assigned as FL users to update model gradients for federated averaging.}
		\label{SemiFL_framework}
		\vspace{-5 mm}
	\end{figure}
	
	\vspace{-3 mm}
	\section{Framework Design}
	As illustrated in Fig. \ref{SemiFL_framework}, we propose a SemiFL framework, where local devices are classified into computing-limited and computing-sufficient devices according to their hardware capabilities \cite{Ahmet2021Hybrid}.
	The former are designated as CL users to upload raw data for centralized training, while the latter are designated as FL users to update model gradients for federated averaging.
	The edge platform consisting of one base station and an edge server is in charge of updating a global model by using the raw data collected from CL users and the local gradients computed by FL users \cite{Ni2022Integrating}.
	By harmonizing federated averaging and centralized training, our SemiFL framework has the following benefits:
	\rmnum{1}) all edge devices in IoT networks are capable of participating in the learning process regardless of their computing capabilities, which enhances the usability of distributed data.
	\rmnum{2}) The huge amount of data gathered from CL users can be used to reconstruct a learning-friendly dataset, even if the statistical characteristics of data are heterogeneous across devices.
	\rmnum{3}) The resource utilization of edge networks can be improved significantly in terms of computation and storage, due to the transmission of datasets and the collaborative computing between local devices and the edge platform.
	
	\begin{figure}[tp]
		\setlength{\abovecaptionskip}{-0.5 mm}
		\centering
		\includegraphics[width= 3.0 in]{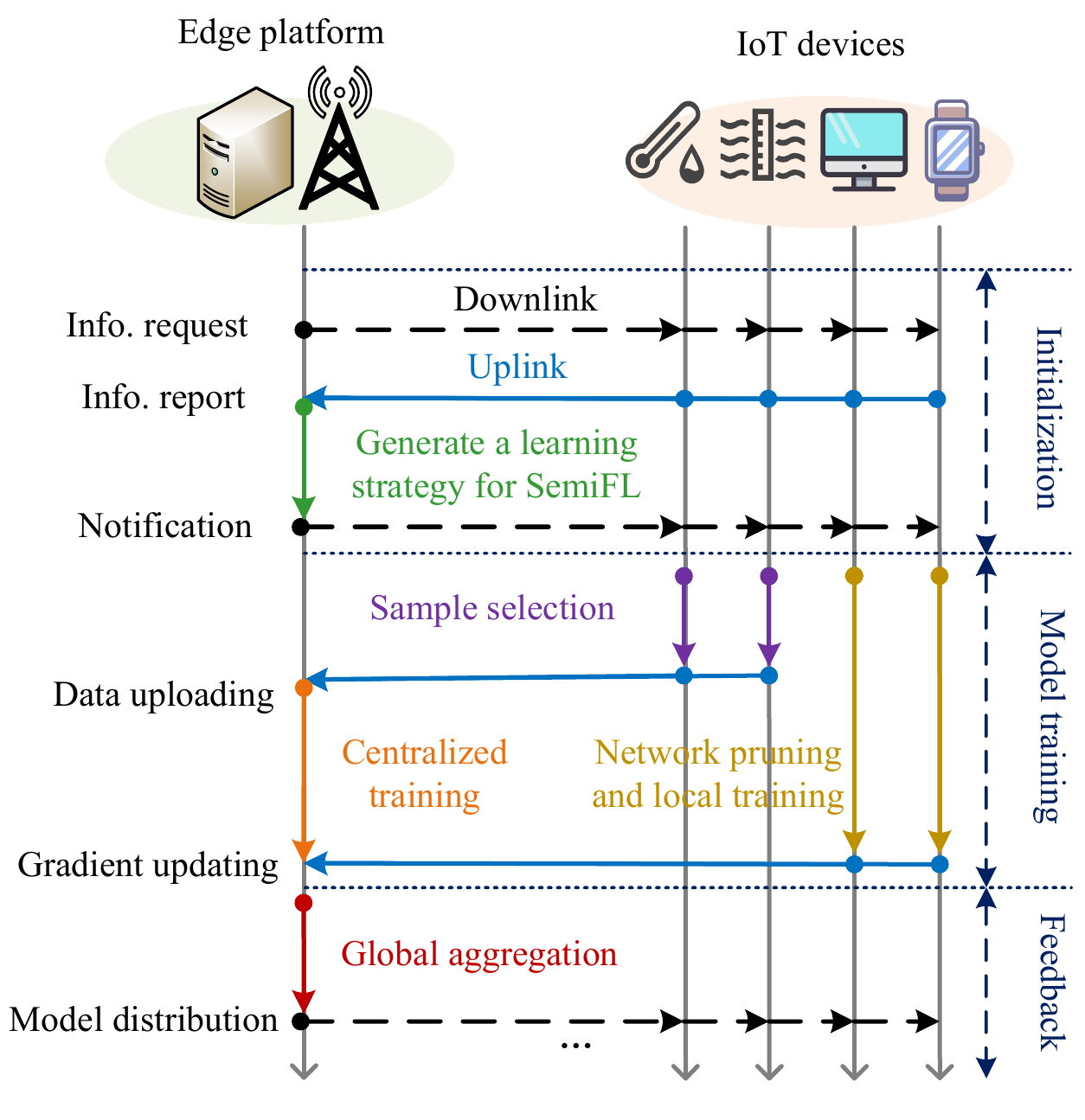}
		\caption{The learning procedure of the proposed SemiFL framework is composed of three stages in one round: initialization, model training, and feedback. The black dashed lines (blue solid lines) denote downlink (uplink) communications between the edge platform and IoT devices.}
		\label{SemiFL_learning_procedure}
		\vspace{-5 mm}
	\end{figure}
	
	As shown in Fig. \ref{SemiFL_learning_procedure}, the overall learning procedure of SemiFL includes three main stages: initialization, model training, and feedback.
	\begin{itemize}
		\item
		\textbf{Stage 1 (Initialization)}:
		The edge platform first requests all devices to report their data amount and distributions, channel conditions, computational resources, etc.
		Based on the received information, the edge platform generates a learning strategy, which is notified to all devices for the subsequent model training stage.
		The basic components of the learning strategy include device classification \cite{Ahmet2021Hybrid}, sample selection \cite{Li2021Sample}, and network pruning \cite{Ren2022Towards}.

		\item
		\textbf{Stage 2 (Model training)}:
		According to the collaborative learning strategy, computing-limited CL users having top $N$ best channels upload partial data samples to the edge platform for centralized training on their behalf \cite{Ahmet2021Hybrid}.
		Meanwhile, all FL users conduct network pruning and local training, then these having top $K$ best models update their gradients to the edge platform \cite{Amiri2021Convergence}.
		
		\item
		\textbf{Stage 3 (Feedback)}:
		Using the gradients obtained by centralized training and federated averaging, the edge platform performs global model aggregation, and then distributes the updated model back to all devices for the next round.
	\end{itemize}
	
	\begin{figure}[tp]
		\setlength{\abovecaptionskip}{-0.1 mm}
		\centering
		\includegraphics[width= 3.0 in]{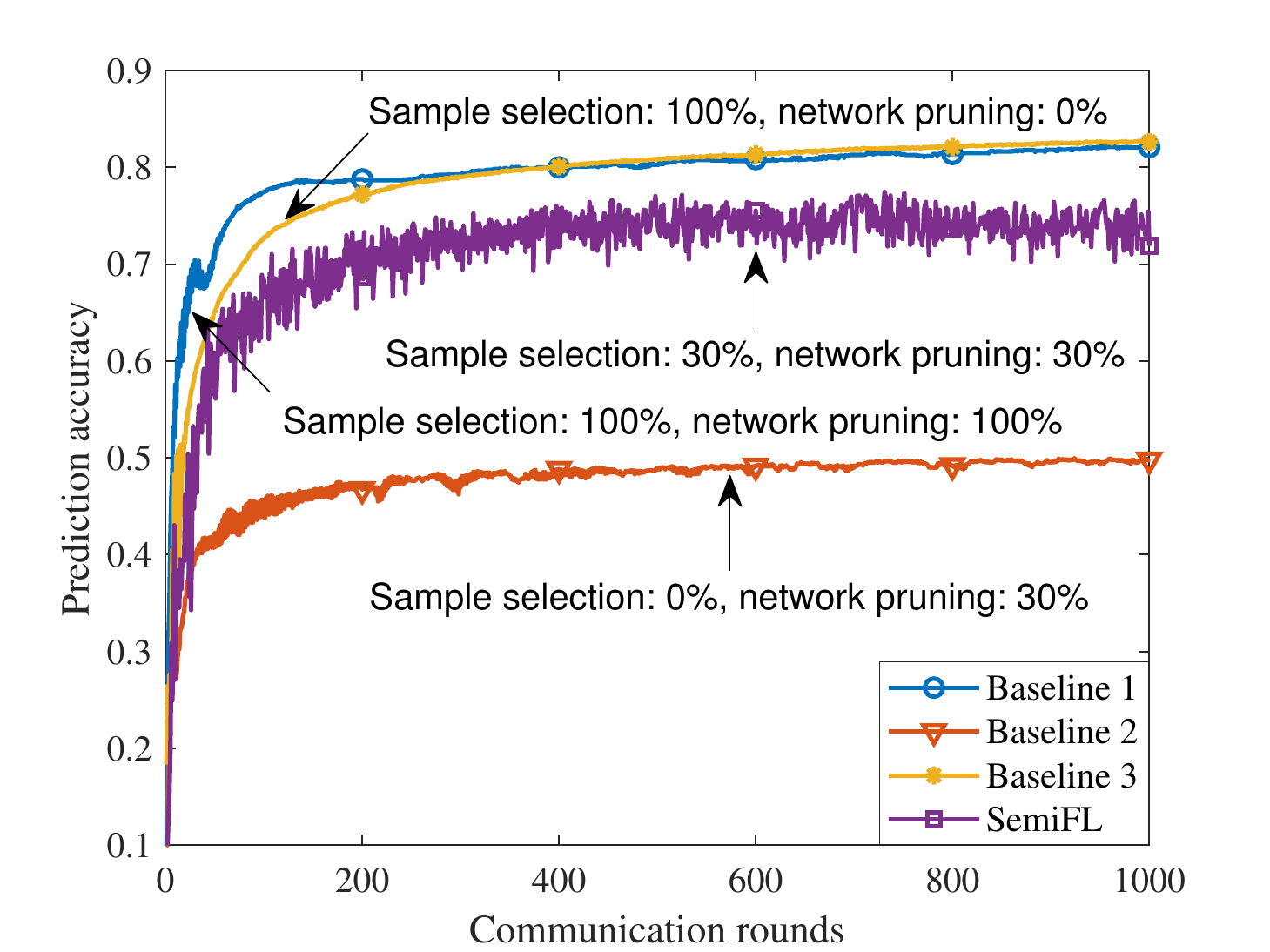}
		\caption{Learning performance on the Fashion-MNIST dataset.}
		\label{learning_performance}
		\vspace{-5 mm}
	\end{figure}
	
	\vspace{-3 mm}
	\section{Experimental Results}
	In the experiments, we consider an IoT network with $4$ computing-limited devices and $6$ computing-sufficient devices.
	We evaluate learning performance of our SemiFL framework by training a multi-layer perceptron on the Fashion-MNIST dataset.
	For comparison, we consider the following baselines.
	\textit{a) Baseline 1 (CL):}
	All devices send their datasets to the edge platform for centralized model training.
	\textit{b) Baseline 2 (FL):}
	All computing-sufficient devices train the model in a federated way.
	\textit{c) Baseline 3 (SemiFL without sample selection and network pruning):}
	All CL users upload their total datasets and all FL users updates their entire models to the server \cite{Ni2022STAR}.
	
	Fig. \ref{learning_performance} demonstrates the learning performance achieved by different benchmarks under non-IID settings.
	Two key observations can be drawn from this figure:
	i) The proposed SemiFL framework without sample selection and network pruning (i.e., \textit{Baseline 3}) outperforms FL (i.e., \textit{Baseline 2}), and achieves the similar performance to CL (i.e., \textit{Baseline 1}) which has higher communication overhead due to the transmission of all local datasets.
	ii) Although sample selection and network pruning can further reduce the communication overhead of SemiFL, they also degrade the achievable prediction accuracy.
	More simulation results and analysis can be found at {\tt https://github.com/niwanli/SemiFL\_IoT}.
	
	\vspace{-3 mm}
	\section{Conclusion}
	A novel SemiFL concept that integrates CL and FL into a harmonized framework has been proposed for IoT networks.
	By allowing the asynchronous data uploading and partial model updating, the proposed SemiFL is capable of mitigating problems of both the data and device heterogeneities that hamper the learning performance in massive IoT networks.
	Besides, due to the change of the computing architecture (from one-side computing to double-side computing), the resources distributed at the network edge can be more fully utilized as compared to conventional CL and FL paradigms.
	In future work, we aim to develop efficient solutions for alleviating the privacy and security issues faced by data-uploading CL users.

\vspace{-3 mm}
\bibliographystyle{IEEEtran}
\bibliography{IEEEabrv,ref}

\end{document}